\documentclass{article}
\usepackage[top=1in, bottom=1in, left=1.5in, right=1.5in]{geometry}
\usepackage[utf8]{inputenc}
\usepackage{graphicx}

\title{\textbf{Activation Functions: Do They Represent A Trade-Off Between Modular Nature of Neural Networks And Task Performance}}

\author{Himanshu Pradeep Aswani, Amit Sethi \\ Indian Institute of Technology, Bombay}

\date{}
\begin{document}
\maketitle
\begin{center}
\section*{Abstract}
Current research suggests that the key factors in designing neural network architectures involve choosing number of filters for every convolution layer, number of hidden neurons for every fully connected layer, dropout and pruning. The default activation function in most cases is the ReLU, as it has empirically shown faster training convergence. We explore whether ReLU is the best choice if one is aiming to desire better modularity structure within a neural network.
\end{center}
\section{Introduction}
Deep learning has been quite successful in achieving great performance on particular tasks, which majorly include, image classification, image segmentation and object detection. The common theme in all of these is the training paradigm of procuring a dataset and either, training a neural network architecture from scratch or fine-tuning a trained network on another task, otherwise known as transfer learning. This approach takes up a lot of computational resources. If, instead, one is to go about developing task-specific modules and come up with meaningful principles to combine them and achieve similar results on tasks currently in focus, it would be a big step in terms of making neural network architectures more reusable as well as serve as to make more efficient the development of even more complicated networks.
\newline
\newline
In this paper, we discuss two different approaches broadly taken to tackle the task of modular nature and explore further in detail, a couple of factors that cause neural networks to depict community structure, the prominent one among them being the choice of activation function.
\section{Related Work}
There has been quite a bit of work done in studying inherent modularity in or designing function specific modules for neural networks. Most of the work done so far could broadly be classified into two classes of thought. These are module composition and module identification which we describe in detail below.
\subsection{Module Composition}
Here, one attempts to explicitly define the functionality of the modules and comes up with methodologies to combine them to achieve a particular task, whether it be image classification, image segmentation or object detection.
For example, [1] explores the task of K-class classification by first constructing modules which are able to distinguish each individual class from the remaining classes and then combines them in a meaningful manner to achieve K-class classification. [2] also explicitly constructs blocks of neurons, which could be viewed as inserting modules for new tasks being incorporated while not forgetting earlier ones. As our understanding of how neural networks work, in deeper layers is still unclear to a certain degree and as this approach demands developing precise modules for specific tasks, it is quite difficult to always execute this principle for a variety of domains with varying shapes of input and output.
\subsection{Module Identification}
Attempting to identify modules that perform particular tasks within a larger neural network and then reusing them elsewhere is the primary goal of this line of thought. [3] is one of the key papers that forms the basis of this type in current literature. It explores a mathematical formulation of identifying clusters within a neural network, factors that affect the extent of cluster within them and information theoretic equivalents of clusters.
\section{Our Method}
Our approach explores the choice of activation function to be used that [3] does not address explicitly. We follow the same graph-theoretic definition of measuring modular nature within a network, as described in [3], by implementing a spectral graph clustering algorithm divide the network into a number of clusters and then measuring the 'goodness measure' of this particular partition. This definition has been motivated by [4]. It requires us to adopt the view of a neural network as an undirected weighted graph. The algorithm mentioned above is utilised to detect community structure within the network.  
\newline
\newline
Before we explain the metric employed, we need to layout some basic definitions. The neural network is viewed as a graph G with all neurons, including input and output neurons, numbered from 1 to N, where N denotes the total number of neurons in the network. We first construct the adjacency matrix A for G by setting A\textsubscript{\textit{i},\textit{j}} = A\textsubscript{\textit{j},\textit{i}} = absolute value of the weight of the edge connecting neurons \textit{i} and \textit{j}. Note that edges in this context occur only between neurons in adjacent layers. If there isn't an edge, we set the value of the weight to 0. The biases are not included under this scheme. The degree d\textsubscript{\textit{i}} of a neuron \textit{i} is defined as $\Sigma$\textsubscript{\textit{j}} A\textsubscript{\textit{i},\textit{j}}.
\newline
\newline
We shall be partitioning the graph into sets of disjoint clusters. For each cluster \textit{Z}, we define vol(\textit{Z}) = $\Sigma$\textsubscript{\textit{i}$\epsilon$\textit{Z}} d\textsubscript{\textit{i}}. For a pair of clusters, \textit{X} and \textit{Y}, we define the weight between them as W(\textit{X},\textit{Y}) =  $\Sigma$\textsubscript{\textit{i}$\epsilon$\textit{X}, \textit{j}$\epsilon$\textit{Y}}A\textsubscript{\textit{i},\textit{j}}. We denote the complement of a cluster \textit{X} as $\bar{X}$, that is the set of all neurons that are not in X. Our metric for a partition with clusters \textit{X}\textsubscript{\textit{1}},...,\textit{X}\textsubscript{\textit{k}} is now defined as ncut(\textit{X}\textsubscript{\textit{1}},...,\textit{X}\textsubscript{\textit{k}}) = $\Sigma$\textsubscript{\textit{i}} W(\textit{X}\textsubscript{\textit{i}},$\bar{X}$\textsubscript{\textit{i}})/vol(\textit{X}\textsubscript{\textit{i}}).
\newline
\newline
The ncut of a neural network is defined as the ncut value of the set of clusters that algorithm 1 returns. A lower ncut value represents a higher extent to which a neural network is modular, and vice versa. To keep computations neither too trivial neither too complicated, we shall be clustering the network into four clusters (k = 4) in the algorithm, the same scheme is employed by [3].
\newline
\newline
We have formulated two methods to explore the effect of choice of activation function to modular nature of networks. The methods differ in the choice of representation of edge weights used as input to the algorithm.
\begin{figure}[htbp]
\centerline{\includegraphics[scale=.5]{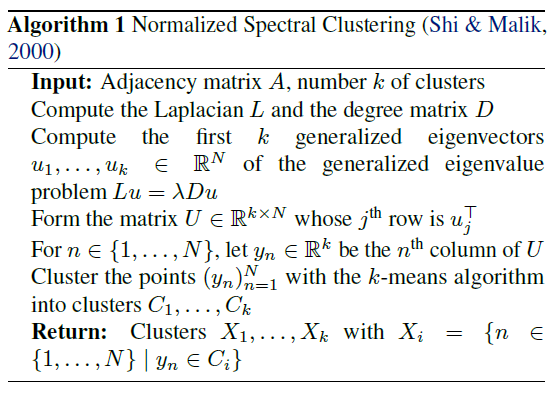}}
\caption{Algorithm 1 reproduced from [3]}
\label{fig}
\end{figure}
\subsection{Method 1: Trained Weights}
The trained weights of the neural network are used as is for this algorithm. We implement an algorithm as described in [4] which is used for the purpose of minimizing the ncut of a graph. [3] has implemented the ncut algorithm for the purpose of module detection in MLPs and explored factors such as dropout, optimization algorithm and network topology. We implement their algorithm but modify their experiments by varying the activation function of the neural network. The popular activation function, ReLU, has a non-differentiable point at zero. Activation functions such Sigmoid do not have non-differential points in their domain. The ncut algorithm is sensitive to weight values and as Sigmoid inherently allows for more non-zero activations at the end of the training phase, we suspect that it will lead to significant changes in the ncut values obtained thus far, using ReLU.
\subsection{Method 2: A Correlation Replacement for Trained Weights}
In this approach, we explore the possibility of high value edge weights distracting the ncut computation in spite of the possibility of pruning it without loss in performance. This situation might occur when the originating node of the high value edge weight itself is rarely activated(this is quite possible when ReLU is used as the activation function). Thus, we propose to replace the value of each edge weight by a correlation measure. The Spearman correlation metric is used to assess the measure of monotonic relation between two random variables in probability theory. We therefore use this measure to compute our correlation measure. Specifically, once the network has been trained, we compute the Spearman correlation measure for every pair of nodes in the network that have an edge between them. To compute this, for every image in the test dataset, the activation of each and every neuron in the network is recorded. For the neurons at the end of the edge under consideration, the Spearman correlation is measured by computing it on the corresponding activation vectors of the neurons which serve as instances of random variables. Once we have computed this for every edge in the network, we run the ncut algorithm on this modified network to gain a sense of the extent of the modular nature in the network.
\section{Experiment}
In this section, we experiment our proposed methods on MLPs trained on the task of image classification on the  MNIST and FashionMNIST datasets. Each dataset has been used as is, both consisting of 60000 training images and 10000 test images. For each dataset, we train four MLPs, differing in choice of activation function and whether dropout has been incorporated or not. Wherever incorporated, the rate of dropout has been set to 0.5. Thus, a total of 8 different MLPs have been studied. All our experiments have been run in PyTorch.
\subsection{Neural Network Architecture}
For each experiment, an MLP has been designed with 4 hidden layers, each of width 256 neurons. The network has been trained for 20 epochs with a batch size of 128 and optimized using the Adam optimization algorithm with default parameters.
\subsection{Method 1 Results}
As described in section 3.1, we apply the ncut algorithm to the weights of the trained MLP. Table 1 shows our results. We observe that, irrespective of the choice of activation function, dropout does lead it to a lower value of ncut. The choice of activation function does affect the test accuracy slightly but introduces a sharper change in ncut values for the MLP. 
\begin{table}
\centering
\begin{tabular}{|c|c|c|c|c|}
     \hline
     \textbf{Data Set} & \textbf{Activation Function} & \textbf{Dropout} & \textbf{Test Accuracy(\%)} & \textbf{N-Cut}\\
     \hline
     MNIST & ReLU & No & 98 & 2.37 \\
     \hline
     MNIST & Sigmoid & No & 97 & 2.10\\
     \hline
     MNIST & ReLU & Yes & 96 & 2.19\\
     \hline
     MNIST & Sigmoid & Yes & 96 & 2.13\\
     \hline
     FashionMNIST & ReLU & No & 88 & 2.13\\
     \hline
     FashionMNIST & Sigmoid & No & 87 &  1.95\\
     \hline
     FashionMNIST & ReLU & Yes & 85 & 2.11\\
     \hline
     FashionMNIST & Sigmoid & Yes & 85 & 1.90\\
     \hline
\end{tabular}
\caption{The table depicts the variation in N-Cut values with activation function where absolute value of edge weight is used to construct the adjacency matrix entries.}
\label{Table 1}
\end{table}
\newline
\subsection{Method 2 Results}
The ncut algorithm has been applied to the neural network with each edge weight replaced by the corresponding Spearman correlation. The results are enumerated in Table 2. Once again, we observe consistent results with those obtained in Table 1 as well as [3]. Dropout does lead to lower ncut value and the choice of activation function does, independently, introduce change in ncut values.
\begin{table}
\centering
\begin{tabular}{|c|c|c|c|c|}
     \hline
     \textbf{Data Set} & \textbf{Activation Function} & \textbf{Dropout} & \textbf{Test Accuracy(\%)} & \textbf{N-Cut}\\
     \hline
     MNIST & ReLU & No & 98 & 2.23 \\
     \hline
     MNIST & Sigmoid & No & 97 & 1.97\\
     \hline
     MNIST & ReLU & Yes & 96 & 2.07\\
     \hline
     MNIST & Sigmoid & Yes & 96 & 1.89\\
     \hline
     FashionMNIST & ReLU & No & 88 & 1.96\\
     \hline
     FashionMNIST & Sigmoid & No & 87 &  1.84\\
     \hline
     FashionMNIST & ReLU & Yes & 85 & 2.01\\
     \hline
     FashionMNIST & Sigmoid & Yes & 85 & 1.65\\
     \hline
\end{tabular}
\caption{The table depicts the variation in N-Cut values with activation function when the Spearman correlation metric between activations of pairs of adjacent neurons is used.}
\label{Table 2}
\end{table}
\section{Conclusion}
The results obtained quite strongly suggest the possibility that by keeping training hyper parameters as well as neural network architecture constant, the choice of activation function may result in a lower learning capability but a larger depiction of modularity within the structure. The reason for this could be attributed to the fact that during backpropogation, the ReLU provides only two distinct derivative values, namely, 0 and 1, whereas the Sigmoid provides all values between 0 to 1 as a possible derivative. This will cause weights to change much more gradually in the case of Sigmoid which influences the nature of the graph obtained.
%As depicted by the results of our experiments, it follows that if we fix the number of training epochs and the architecture for a particular dataset, the sigmoid activation function results in a marginally less accurate neural network with a significant extent of clusterable nature or community structure creeping into the model when compared to the relu being chosen. This can be attributed, as we had hypothesized above, to the nature of the sigmoid function as compared to the relu function. This is one example that seems to suggest that always choosing the relu is not always better as if you want a modular nature in the network, a different activation function may be able to do just that, while slightly sacrificing on accuracy. Our work suggests a trade-off which was not earlier apparent in the design of neural network architectures. Most of the major tasks rely on the relu activation function as it provides faster training convergence as well as higher accuracy without any seemingly obvious consequences. We demonstrate that a trade-off does occur in terms of the extent of community structure developed in a network during training time.
\newline
\newline
One of the key questions that need to be answered now is how does one exlpoit this modularity structure within a neural network. The relation between modularity structure and task-specific sub-modules in a neural network isn't quite apparent at this stage. We hope that future research will help shed light on this matter as well as advocate the notion of sub-module identification as another way to achieve better neural network re-usability without the need of fine-tuning it for a new task.
\section*{References}
1. Bao-Liang Lu and Masami Ito Task Decomposition and Module Combination Based on Class Relations: A Modular Neural Network for Pattern Classification.
\newline
\textit{IEEE transactions on neural networks, Vol. 10, No. 5, September 1999}
\newline
2. Alexander V. Terekhov, Guglielmo Montone, J. Kevin O'Regan Knowledge transfer in deep block-modular neural networks. \textit{arXiv:1908.08017 [cs.NE]}
\newline
3. Daniel Filan, Shlomi Hod, Cody Wild, Andrew Critch, Stuart Russell Neural Networks Are Surprisingly Modular. \textit{arXiv:2003.04881 [cs.NE]}
\newline
4. Shi, J. and Malik, J. Normalized cuts and image segmentation.
\textit{IEEE Transactions on pattern analysis and machine
intelligence, 22(8):888–905, 2000.}
\newline
5. Newman, M. E. Modularity and community structure in networks.
\textit{Proceedings of the national academy of sciences,
103(23):8577–8582, 2006.}
\end{document}